\documentclass[journal]{IEEEtran}

\usepackage{xcolor,soul,framed} 

\colorlet{shadecolor}{yellow}
\usepackage[pdftex]{graphicx}
\graphicspath{{../pdf/}{../eps/}}
\DeclareGraphicsExtensions{.pdf,.eps,.png}

\usepackage[cmex10]{amsmath}
\usepackage{amssymb}
\usepackage{mathabx}
\usepackage{amsthm,amsmath,amssymb}
\usepackage{caption}
\usepackage{array}
\usepackage{mdwmath}
\usepackage{mdwtab}
\usepackage{eqparbox}
\usepackage{url}
\usepackage{cite}
\usepackage{booktabs}
\usepackage{multirow}
\usepackage{bm}
\usepackage{mathrsfs}
\usepackage{algorithmic}
\usepackage[ruled]{algorithm2e}
\usepackage{setspace}

\begin{document}
\bstctlcite{IEEEexample:BSTcontrol}
    \title{Improving Adversarial Robustness of Deep Neural Networks by Using Semantic Information}
  \author{Lina~Wang,~Xingshu~Chen,~Rui~Tang,~Yawei~Yue,~Yi~Zhu,~Xuemei~Zeng,~Wei~Wang

  \thanks{L.~Wang,~X.~Chen,~R.~Tang~and~Y.~Yue~ are with the School of Cyber Science and Engineering at Sichuan University, Chengdu 610065, China (E-mail:wlnlnw1992@163.com, chenxsh@scu.edu.cn, 2017326240002@stu.scu.edu.cn., yue123161@stu.scu.edu.cn) }
  \thanks{Y.~ZhuX.~Zeng~and~W.~Wang is with the Cyber Science Research Institute at Sichuan University, Chengdu 610065, China (E-mail:zhuyi20@scu.edu.cn, zengxm@scu.edu.cn, wwzqbx@hotmail.com).}}%

\markboth{Knowledge-Based Systems 226 (2021) 107141
}{Wang \MakeLowercase{\textit{et al.}}: Improving Robustness by Semantic Information}

\maketitle

\begin{abstract}
The vulnerability of deep neural networks (DNNs) to adversarial attack, which is an attack that can mislead state-of-the-art classifiers into making an incorrect classification with high confidence by deliberately perturbing the original inputs, raises concerns about the robustness of DNNs to such attacks. Adversarial training, which is the main heuristic method for improving adversarial robustness and the first line of defense against adversarial attacks, requires many sample-by-sample calculations to increase training size and is usually insufficiently strong for an entire network. This paper provides a new perspective on the issue of adversarial robustness, one that shifts the focus from the network as a whole to the critical part of the region close to the decision boundary corresponding to a given class. From this perspective, we propose a method to generate a single but image-agnostic adversarial perturbation that carries the semantic information implying the directions to the fragile parts on the decision boundary and causes inputs to be misclassified as a specified target. We call the adversarial training based on such perturbations ``region adversarial training'' (RAT), which resembles classical adversarial training but is distinguished in that it reinforces the semantic information missing in the relevant regions. Experimental results on the MNIST and CIFAR-10 datasets show that this approach greatly improves adversarial robustness even when a very small dataset from the training data is used; moreover, it can defend against fast gradient sign method, universal perturbation, projected gradient descent, and Carlini and Wagner adversarial attacks, which have a completely different pattern from those encountered by the model during retraining.
\end{abstract}


%
\IEEEpeerreviewmaketitle


\section{Introduction}
\label{section:1}

\IEEEPARstart{A}{s} an accepted technique in machine learning, deep learning (DL) has proved itself capable of performing singularly well on a number of categories of machine learning tasks \cite{Lecun2015Deep}. In particular, deep neural networks (DNNs) can learn very effective models for input classification. State-of-the-art DNNs have achieved impressive performance in tasks of computer vision \cite{krizhevsky2009learning}, \cite{Francisco2020Object}, speech recognition \cite{hinton2012deep}, \cite{G2019Posterior}, and natural language understanding \cite{sutskever2014sequence}, \cite{Basemah2020Improving} and provide solutions based on these tasks for many other problems, such as in medical science \cite{esteva2017dermatologist}. The universal approximator theorem \cite{hornik1989multilayer} guarantees the representational power of DNNs but does not indicate whether a training algorithm will be able to discover a function having all the desired properties.

\begin{figure}
    \centering
    \includegraphics[width=3.2in]{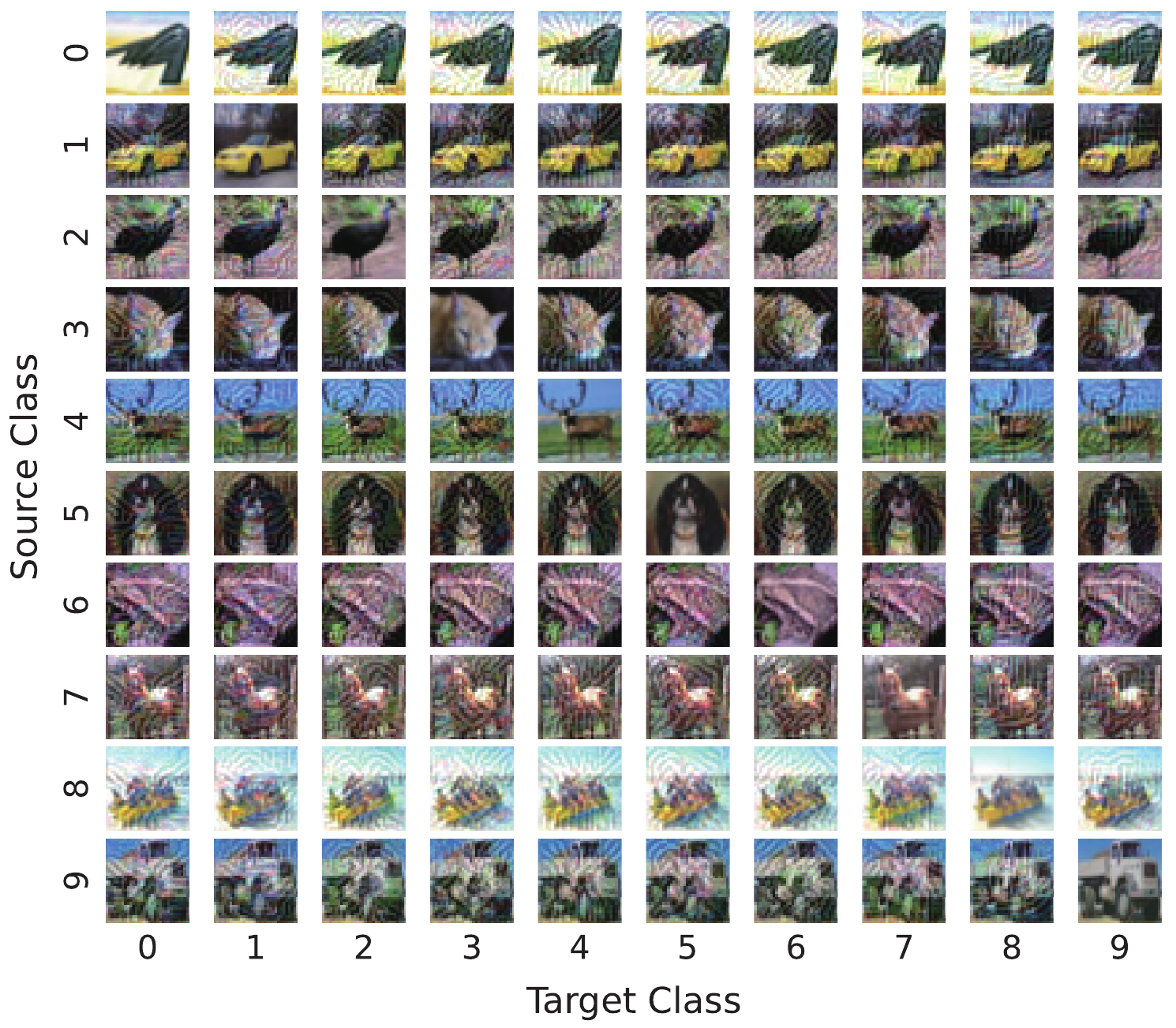}
    \caption{Illustration of targeted universal perturbations (TUPs) attacks on a typical DNN using images sampled from CIFAR-10, showing source--target pairs. To facilitate the presentation, we use the numbers 0--9 to represent the ten classes of CIFAR-10. The number preceding each row represents the source class, and the number preceding each column represents the target class. For example, an image with a row number of 1 and a column number of 2 is a TUP adversarial example whose true label is class 1 but is incorrectly classified as class 2. All of the original images displayed were selected at random.}
    \label{fig:adversarial}
\end{figure}

For all the success of deep learning algorithms, Szegedy et al. \cite{szegedy2013intriguing}, \cite{goodfellow2014explaining} revealed an inherent weakness of DNNs by pointing out the existence of a new type of attack called an adversarial attack. The adversary in this type of attack misleads models into producing an incorrect output with an adversarial example, a plausible member of input datasets that is only slightly different from benign examples, created by adding a carefully constructed adversarial perturbation. For example, the images on the diagonal in Fig.~\ref{fig:adversarial} are unperturbed clean examples, and the other images are adversarial examples misclassified as specified target classes that are almost imperceptible to human vision.  Recent studies have made it clear that DNNs are universally vulnerable to adversarial examples; this seems to contradict the assumptions that underlie many deep learning methods and suggests that our deep classifiers based on modern machine learning techniques have only built a Potemkin village instead of learning the true underlying concepts that determine a correct output label.
Ideally, the label estimated by a classifier should not be altered by a sufficiently small perturbation of an input data point, let alone an adversarial perturbation. This excellent property, called robustness, is extremely significant for DNNs when applied in realistic contexts, and above all in security-critical environments \cite{Liu2018Improving}. Because of the importance and imminence of the issue, the robustness of classifiers to adversarial examples has been attracting much attention in recent years.

Previous studies on the robustness of DNNs have approached the question from two directions, attempting either to prove a lower bound of robustness through formal guarantees or to find an upper bound of robustness through adversarial attacks. The formal approach is sound but difficult to carry out in practice \cite{raghunathan2018certified}, whereas heuristic defenses against adversarial attacks are not sufficiently strong \cite{carlini2017towards}. There is a puzzling problem concerning the latter approach. It is generally believed that neural networks are not learning the true concepts \cite{goodfellow2014explaining}, yet the adversarial perturbations generated by almost all known methods appear to be chaotic! This seems counter-intuitive, because if the network is missing important information related to the true underlying concepts, this information should be reflected in the adversarial examples, representing the blind spots of the network.

In addition, despite a number of meaningful studies on the issue, achieving ideal robustness remains a difficult goal. Improving the adversarial robustness of a network as a whole is rather ambitious and difficult; sometimes enhancing the robustness of particular regions in the manifold represented by the network can provide a greater benefit in reality. This is even more remarkable for certain application scenarios, especially security-sensitive applications. For example, for a classifier that distinguishes different kinds of animals, it is no more dangerous to classify dogs as cats than dogs as birds, but in the case of a multi-category classifier for malware classification or for traffic sign recognition as used in autonomous vehicles, things are quite different \cite{stallkamp2012man}. Incorrectly classifying a yield sign as a stop sign is likely to be safer than misclassifying it as a sign that allows vehicles to pass. Similarly, misclassifying malware as belonging to the wrong malware family is less harmful than incorrectly classifying it as benign.

Furthermore, almost all heuristic methods for improving adversarial robustness require a large number of calculations on a very large dataset of a size comparable to that of the training set. This considerably reduces their suitability for practical scenarios, especially application environments having high timeliness requirements.

In this paper, we focus on the region corresponding to a certain class in the manifold represented by the attacked network, and we propose a method to extract semantic information that is universal for most examples from a small set of the data points that lie very close to the classifier's decision boundary separating one class from all others. The key idea is to emphasize to the classifier the semantic information it has not yet learned and to prompt the classifier to learn a clearer (usually more complicated) decision boundary and the underlying concepts. We retain this universality property across the inputs as \cite{moosavi2017universal} did, but unlike researchers in previous studies, we generate perturbations containing semantic information instead of meaningless noise with the aim of improving robustness.
The main contributions of this paper are as follows:

\begin{itemize}
    \item We find that there exists a single perturbation applicable to most of the inputs that could constitute a targeted adversarial attack on a classifier, and, importantly, that such perturbations are not meaningless but contain explicit semantic information.
Furthermore, we propose an algorithm for generating such targeted universal perturbations (TUPs). The algorithm computes a series of perturbation vectors one at a time, sending a data point to the classification boundary of the region corresponding to the specified target class for a set of points in the training dataset, and then aggregates the perturbation vectors to find a universal vector indicating the direction to the region in an iterative way. We show that the proposed algorithm can calculate such a perturbation on a very small set of training data points, which causes new samples to be misclassified as a specific target class with high probability.
    \item  We present a new approach to improve adversarial robustness, called region adversarial training (RAT), and formalize it conceptually. RAT pays special attention to the region near the decision boundary corresponding to a selected target class and then uses the extracted semantic information related to this region to guide the retraining process. The information used by RAT comprises the common patterns for most samples that follow the same distribution as the training data, and these patterns contain semantic information related to the true underlying concepts; consequently, RAT can not only perform well on a very small data set, but also defend against adversarial attacks that have never been seen by the network before.
    \item We validate the algorithm by reporting the results of extensive experiments using MNIST \cite{lecun1998mnist} and CIFAR-10 \cite{krizhevsky2009learning} and show that the perturbations achieve a similar high attack success rate for each target class. We also systematically evaluate the choice of algorithm parameters. We find that our TUP perturbations not only retain the universality property of being able to fool unseen data points but also transfer well across different architectures and can work well even when calculated from a very small dataset. We experimentally demonstrate that when the proposed algorithm is employed to provide examples for region adversarial training (even on a very small set from training data), the test set accuracy on both TUP adversarial examples and the best-known Carlini and Wagner (C\&W) \cite{carlini2017towards},  universal perturbation (Uni.) \cite{moosavi2017universal}, projected gradient descent (PGD) \cite{madry2017towards}, fast gradient sign method (FGSM) \cite{goodfellow2014explaining}  adversarial examples can be increased on MNIST and CIFAR-10.
\end{itemize}

The rest of this paper is organized as follows. In Section~\ref{section:2}, we summarize recent work on generating adversarial examples and improving adversarial robustness. Section~\ref{section:3} provides the preliminaries and defines the notation. Then, we introduce the proposed approaches for finding TUPs and formalize the region adversarial training method in Section~\ref{section:4}. The experiments we conducted to test the proposed method are described and their results analyzed in Section~\ref{section:5}. Finally, we conclude with Section~\ref{section:6}.

\begin{figure*} 
    \centering
    \includegraphics[width=\textwidth]{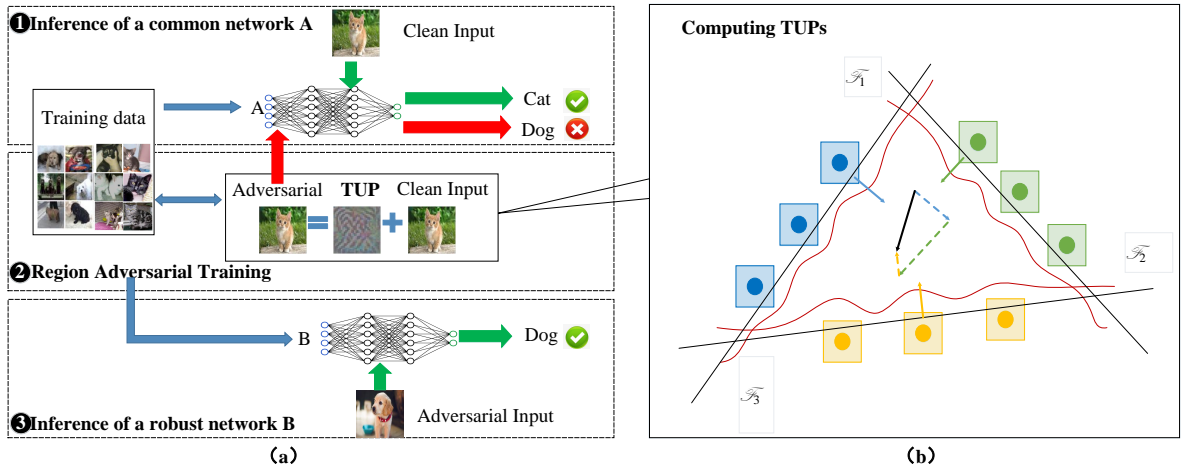}
    \caption{Overview of region adversarial training (RAT) based on TUP adversarial examples. (a) shows the entire RAT process. (b) is an illustration of Algorithm~\ref{alg:1} for computing a TUP perturbation.}
    \label{fig:idea}
\end{figure*}

\section{Related Work}
\label{section:2}
As our goal is to extract missed semantic information through a method of generating adversarial examples and then to improve the robustness of DNNs, this section first introduces the work related to the generation of adversarial examples and then describes the studies on improving adversarial robustness.

\textbf{Adversarial examples.} Szegedy et al. \cite{szegedy2013intriguing} discovered the existence of the possibility of adversarial attacks on deep neural networks by generating adversarial examples using box-constrained L-BFGS. The fact that deep neural networks are surprisingly susceptible to such adversarial attacks triggered the wide interest of researchers in the security and machine learning communities, and since then, a sizable body of related literature has introduced several new methods for crafting adversarial examples to construct an upper bound on the robustness of neural networks. Goodfellow et al. \cite{goodfellow2014explaining} proposed a method called ``Fast Gradient Sign Method'' (FGSM), which perturbs an image to increase the loss of the classifier on the resulting image based on the ``linearity hypothesis'' of deep network models in higher-dimensionality space. Instead of using the $L_2$-norm as in FGSM, Kurakin et al. \cite{kurakin2016adversarial} presented an alternative approach named ``Fast Gradient $L_\infty$'' and also extended FGSM to a ``target class'' variation wherein the label of the class least likely to be predicted by the attacked network is used as the target class. Unlike the one-step methods, which take a single step in the direction that increases the loss, the Basic Iterative Method (BIM) \cite{kurakin2016adver} computes the perturbation iteratively by adjusting the direction step by step. Papernot et al. \cite{papernot2016limitations} modified pixels of the original image one at a time by computing a saliency map and then monitored the effect of the changes.  A more refined algorithm, named DeepFool \cite{moosavi2016deepfool}, moves a given image toward the boundary of a polyhedron through a small vector based on an iterative linearization of the classifier to compute a minimal-norm adversarial perturbation. C\&W \cite{carlini2017towards} introduced a set of three adversarial attacks in the wake of defensive distillation against the adversarial attacks and inspired the ``Zeroth Order Optimization (ZOO)'' attack which was the first optimization-based attack in black-box settings.
The method proposed by Su et al. \cite{su2019one} was deduced for the extreme case in which only one pixel in the image is allowed to change for the attacker, and they reported a fairly good success rate, $70.97\%$. Hybrid Attacks \cite{247692} combined the two strategies, optimization-based attacks and transfer attacks in the black-box setting, to reduce cost and improve success rates. All of the above methods compute adversarial perturbations to fool an attacked network using a single image; the method in \cite{moosavi2017universal} is fundamentally different. The authors computed perturbations that do not involve a data-dependent optimization but fooled the classifier on all images through one and the same perturbation. However, the perturbations they performed caused the clean samples to be misclassified as any (unpredictable) class and contained very little semantic information. By contrast, our approach computes a perturbation that moves the sample in a specific direction chosen to cause the perturbed sample to be misclassified as a target class $t$ while preserving the universality property across samples, without the need to use any complex generative models such as in \cite{sarkar2017upset}. More importantly, our approach extracts explicit semantic information with very few samples and generates adversarial perturbations that show this semantic information clearly and that exhibit a pattern completely different from the others.

\textbf{Adversarial robustness.} The appearance of adversarial examples reveals the intrinsic vulnerability of the existing neural network methodology; therefore, studies on improving its robustness to adversarial examples are of great importance. Work has generally been developing in two different directions. One way of making neural networks robust to adversarial attacks focuses on formally ensuring their robustness. Robustness verification is a general method for obtaining safety guarantees \cite{DBLP:conf/iclr/TjengXT19}, \cite{wang2018formal}, \cite{gehr2018ai2}, \cite{wong2018provable}, \cite{singh2018fast}, \cite{weng2018towards}, \cite{zhang2018efficient}; it is typically based on sophisticated theory and is usually computationally expensive. As the investigation in this paper does not involve formal verification techniques, we do not go into detail here.
The other way is to explore heuristic defenses against adversarial examples (including their detection), by means of modifying networks directly  \cite{gu2014towards}, \cite{rifai2011contractive}, \cite{ross2018improving}, \cite{lyu2015unified}, \cite{nguyen2018learning}, \cite{papernot2016distillation}, \cite{nayebi2017biologically}, \cite{krotov2018dense}, \cite{cisse2017houdini}, \cite{gao2017deepcloak}, \cite{madry2017towards}, \cite{na2017cascade}, using extra network add-ons \cite{akhtar2018defense}, \cite{xu2017feature}, \cite{shen2017ape}, \cite{lee2017generative}, or changing the training procedure or using modified inputs in the inference phase \cite{sankaranarayanan2018regularizing}, \cite{miyato2016adversarial}, \cite{zheng2016improving}, \cite{dziugaite2016study}, \cite{guo2017countering}, \cite{das2017keeping}, \cite{luo2015foveation}, \cite{mopuri2017fast}. The method presented in this paper is of this type and, more specifically, falls into the category of \textit{adversarial training} \cite{goodfellow2014explaining}, \cite{shaham2018understanding}, which modifies the training procedure with adversarial inputs. What most distinguishes our work from other adversarial training methods is that whereas to our knowledge all existing methods improve the adversarial robustness of networks as a whole, ours focuses on certain regions in the manifold represented by the network. In addition, all existing studies on adversarial training have used an image-specific method to increase the size of the training dataset, which requires at least one calculation for each example on a very large dataset (usually a multiple of the training set). To the best of our knowledge, the method in \cite{moosavi2017universal} is the only exception; it calculates a single image-agnostic perturbation for a set of training points, but it leads to only a slight improvement in robustness. Our method is designed to enhance the robustness of DNNs on a very small set by using perturbations that contain semantic information and retain the universality property but that
are completely different from the patterns in \cite{moosavi2017universal}.
\section{Preliminaries}
\label{section:3}
\subsection{Neural networks: Definitions and notation}
A neural network used as a multi-class classifier, which is the case exclusively studied in this paper, is given an input and provides a corresponding class probability vector as output. Formally, a classifier $\widehat{f} \colon R^{n} \rightarrow \{1 \textellipsis K\}$ accepts an input $x \in R^{n}$ and provides an estimated label $\widehat{f}(x)$ as output for it. We assume that $x\sim\psi$, where $\psi$ denotes a distribution of inputs in $R^n$. The output vector $\widehat{f}(x)$ represents the probability that the input $x$ belongs to each of the $K$ classes. The classifier assigns the label $\widehat{y}(x)= \mathop{argmax}\widehat{f}(x)_i$ to the input $x$; the ground-truth label is denoted by $y$. The model $\widehat{f}$ depends on some parameters $\theta$, but as the network is fixed for our method of crafting an adversarial perturbation, we will omit $\theta$ from $\widehat{f}$ when there is no ambiguity. We define $J(\bm{\theta},\bm{x},y)$ as the loss function used to train the model.
\subsection{Adversarial examples}
\label{section:3.2}
Given a naturally occurring example (clean example) $x$ and a classifier $\widehat{f}(\cdot)$, an adversarial example \cite{szegedy2013intriguing} is an input that causes the classifier to make a mistake. An adversary launches adversarial attacks by crafting adversarial examples. Let $x'=x+r$ be an adversarial example that is very similar to $x$, where $r$ is a small vector called an adversarial perturbation. More precisely, an untargeted adversarial example is one that causes the classifier to predict any incorrect label (i.e., it makes $\widehat{f}(x') \neq \widehat{f}(x)$), and a targeted adversarial example is one that causes the classifier to change the prediction to some specific target class $t$ (i.e., $\widehat{f}(x')=t$). It is apparent that untargeted adversarial attacks are strictly less powerful than targeted adversarial attacks, meaning that if an adversarial example can cause a targeted adversarial attack, it can certainly cause an untargeted adversarial attack\cite{kurakin2018adversarial},\cite{akhtar2018threat}. The similarity between $x$ and $x'$ is usually measured by some distance metric $d(\cdot)$. In the literature for generating adversarial examples, the three distance metrics $L_0$-norm, $L_2$-norm, and $L_\infty$-norm (collectively, $L_p$-norms) are widely used. The $L_p$-norm of a vector $v$ is defined as
\[\left \| v \right \|_{p}=\left ( \sum_{i=1}^{n}\left | v_{i} \right |^{p} \right )^{\frac{1}{p}}.\]
In this paper, we focus on the $L_\infty$ distance. It is true that no distance metric is a perfect measure of human perception, especially considering different scenarios. Constructing and evaluating a good distance metric may be intuitive, but we do not judge which distance metric is optimal as it is not the focus of this paper. Instead, we use $L_\infty$  distance, as $L_p$ is sufficient for the computer vision classification task that is the focus of this paper, and $L_\infty$ norm is considered as the optimal choice \cite{carlini2017towards} and has been widely used in many studies \cite{papernot2016effectiveness}\cite{warde201611}.
\subsection{Threat model}
The threat model of a system, which often involves adversarial goals and capabilities, can be used to measure the security of the system. If a system using a DNN is viewed as a generalized data processing pipeline, at inference phase the system collects inputs from sensors or data repositories and then processes the inputs in the digital domain and feeds them to the model to produce an output for external systems or users to receive and act upon. According to the attack surface defined with this procedure, in this paper we consider adversaries that are capable of manipulating the collection and processing of data to tamper with the output. The adversaries have no knowledge of the model architecture or values of any parameters or trainable weights, but they have direct access to at least some of the training data, and of course they can query the model, i.e., feed it inputs and receive outputs. Finally, by modeling the adversarial goals using a classical approach that includes confidentiality, integrity, and availability, called CIA \cite{Guttman1995An}, it can be seen that the main threat from such adversaries is to compromise the integrity of the DNN-based system. As they are capable of destroying the input--output mapping of the model, they can also achieve the goal of undermining availability, despite the difference between availability and integrity in definition.

\section{Preliminaries}
\label{section:3}
\subsection{Neural networks: Definitions and notation}
A neural network used as a multi-class classifier, which is the case exclusively studied in this paper, is given an input and provides a corresponding class probability vector as output. Formally, a classifier $\widehat{f}\colon R^n \rightarrow\{1\textellipsis K\}$ accepts an input $x\in R^n$ and provides an estimated label $\widehat{f}(x)$ as output for it. We assume that $x\sim\psi$, where $\psi$ denotes a distribution of inputs in $R^n$. The output vector $\widehat{f}(x)$ represents the probability that the input $x$ belongs to each of the $K$ classes. The classifier assigns the label $\widehat{y}(x)= \mathop{argmax}\widehat{f}(x)_i$ to the input $x$; the ground-truth label is denoted by $y$. The model $\widehat{f}$ depends on some parameters $\theta$, but as the network is fixed for our method of crafting an adversarial perturbation, we will omit $\theta$ from $\widehat{f}$ when there is no ambiguity. We define $J(\bm{\theta},\bm{x},y)$ as the loss function used to train the model.
\subsection{Adversarial examples}
\label{section:3.2}
Given a naturally occurring example (clean example) $x$ and a classifier $\widehat{f}(\cdot)$, an adversarial example \cite{szegedy2013intriguing} is an input that causes the classifier to make a mistake. An adversary launches adversarial attacks by crafting adversarial examples. Let $x'=x+r$ be an adversarial example that is very similar to $x$, where $r$ is a small vector called an adversarial perturbation. More precisely, an untargeted adversarial example is one that causes the classifier to predict any incorrect label (i.e., it makes $\widehat{f}(x') \neq \widehat{f}(x)$), and a targeted adversarial example is one that causes the classifier to change the prediction to some specific target class $t$ (i.e., $\widehat{f}(x')=t$). It is apparent that untargeted adversarial attacks are strictly less powerful than targeted adversarial attacks, meaning that if an adversarial example can cause a targeted adversarial attack, it can certainly cause an untargeted adversarial attack\cite{kurakin2018adversarial},\cite{akhtar2018threat}. The similarity between $x$ and $x'$ is usually measured by some distance metric $d(\cdot)$. In the literature for generating adversarial examples, the three distance metrics $L_0$-norm, $L_2$-norm, and $L_\infty$-norm (collectively, $L_p$-norms) are widely used. The $L_p$-norm of a vector $v$ is defined as
\[\left \| v \right \|_{p}=\left ( \sum_{i=1}^{n}\left | v_{i} \right |^{p} \right )^{\frac{1}{p}}.\]
In this paper, we focus on the $L_\infty$ distance. It is true that no distance metric is a perfect measure of human perception, especially considering different scenarios. Constructing and evaluating a good distance metric may be intuitive, but we do not judge which distance metric is optimal as it is not the focus of this paper. Instead, we use $L_\infty$  distance, as $L_p$ is sufficient for the computer vision classification task that is the focus of this paper, and $L_\infty$ norm is considered as the optimal choice \cite{carlini2017towards} and has been widely used in many studies \cite{papernot2016effectiveness}\cite{warde201611}.
\subsection{Threat model}
The threat model of a system, which often involves adversarial goals and capabilities, can be used to measure the security of the system. If a system using a DNN is viewed as a generalized data processing pipeline, at inference phase the system collects inputs from sensors or data repositories and then processes the inputs in the digital domain and feeds them to the model to produce an output for external systems or users to receive and act upon. According to the attack surface defined with this procedure, in this paper we consider adversaries that are capable of manipulating the collection and processing of data to tamper with the output. The adversaries have no knowledge of the model architecture or values of any parameters or trainable weights, but they have direct access to at least some of the training data, and of course they can query the model, i.e., feed it inputs and receive outputs. Finally, by modeling the adversarial goals using a classical approach that includes confidentiality, integrity, and availability, called CIA \cite{Guttman1995An}, it can be seen that the main threat from such adversaries is to compromise the integrity of the DNN-based system. As they are capable of destroying the input--output mapping of the model, they can also achieve the goal of undermining availability, despite the difference between availability and integrity in definition.

\label{section:4}
An overview of the method for generating TUP adversarial examples and performing region adversarial training is given in Fig.~\ref{fig:idea}(a). A conceptual illustration of the method for computing TUPs is presented in Fig.~\ref{fig:idea} (b). As shown in Fig.~\ref{fig:idea} (a), a common neural network $A$ can correctly classify a clean input but cannot correctly classify an adversarial example in the inference phase. Retraining using our RAT method based on TUPs results in a more robust network $B$ that can correctly classify even the unseen adversarial examples. In Fig.~\ref{fig:idea} (b), we use black solid lines represent a simple decision boundary (which is linear in this case) for the original network. A set of data points can be easily separated with the simple decision boundary, but the $L_\infty$ balls around the data points cannot be separated well. Let $\mathscr{F}_k = \left \{ x:\hat{f}_k(x)- \hat{f}_t(x)=0 \right \}$ (in the case shown in Fig.~\ref{fig:idea}(b), $k=1, 2, 3$) describe the region of the space where the classifier outputs label $t$. For each point whose ground-truth label is not $t$ but is not classified correctly by the simple decision boundary, the method calculates a vector that touches a polyhedron that approximates the region $\mathscr{F}_k$. Then, by continuously aggregating these vectors and updating the perturbation vector, we finally obtain a TUP perturbation that captures the semantic information that the network has not learned but is about the decision boundary of the region where the classifier outputs label $t$. Using this information to retrain the network, a more complicated decision boundary, needed to separate adversarial examples in the $L_\infty$ balls, can be obtained (represented by the red curve in Fig.~\ref{fig:idea}(b)). This makes the resulting network more robust against adversarial attacks with bounded $L_\infty$ perturbations. Note that, the Algorithm~\ref{alg:1} and the Algorithm~\ref{alg:2} presented below make two assumptions. First of all, our algorithm is applicable to classifiers that satisfy the assumption that the training data and test data are independent and identically distributed (i.i.d). Secondly, we use $L_\infty$ norm to measure the similarity of the examples before and after attack, as mentioned in Section~\ref{section:3.2}. Using different distance metrics will affect the effectiveness of the algorithm.
\subsection{Targeted universal perturbations}
The problem of generating an adversarial example for an input $x$ is equivalent to that of finding a minimum adversarial perturbation $r$ that satisfies the adversarial condition. Formally, this problem can be defined as follows:
\begin{equation}
\begin{split} 
\min_{r}  \quad d(x, x+r)\\
s.t.~\widehat{f}(x + r) = t.
\end{split}
\label{P}
\end{equation}
In Eq.~\eqref{P}, $x$ and $x+r$ must be drawn from the same distribution $\psi$ and the same feature space. As our aim is to cause a targeted adversarial attack for most inputs through a single perturbation and to extract semantic information from them, the problem differs a bit. Our generation method focuses on the following question: Can we find a perturbation vector $r\in R^n$ that causes the classifier to misclassify almost all data points sampled from $\psi$ as a certain class $t$ that differs from the correct prediction for the original input? In other words, we look for a vector $r$ for most $x\sim\psi$ such that
 \begin{equation}
     \widehat{f}(x + r) = t \neq\widehat{f}(x).
 \end{equation}
According to the concepts of adversarial examples and adversarial perturbations as described before, each of the following two constraints on the perturbation vector $r$ must be satisfied:
\begin{subequations}
\begin{align}
d(r)\leq\eta \label{Za}\\
\mathop{P}\limits_{x\sim\psi}(\widehat{f}(x+r)=t)\geq1-\delta. \label{Zb}
\end{align}
\end{subequations}

In Eq.~\eqref{Za}, we use $d(r)$ as a measure of the quantified similarity. In Eq.~\eqref{Zb}, we use $1-\delta$ to denote the success rate threshold, where the parameter $\delta\in (0,1]$ is a scalar.  The parameter $\eta$ restricts the magnitude of the perturbation. The smaller the value of $\eta$, the harder it is for a human to perceive the perturbation in the image, and on the other hand, a larger $1-\delta$ value (i.e., a smaller $\delta$ value) implies a stronger attack that is more powerful for generating a desired perturbation.
We call such a perturbation $r$ a \textit{targeted-$(\delta,\eta)$-universal perturbation} (TUP), as this single input-agnostic perturbation, restricted by the parameters $\delta$ and $\eta$, causes the predicted label of most data points sampled from the data distribution $\psi$ to be converted to the target class $t$.

\textbf{Algorithm 1.} In this paper, we propose an algorithm that seeks a common perturbation $r$ for most data points in $X = \{x_1,\ldots,x_s\}$, which is a set of images sampled from the same distribution $\psi$, such that the attacked neural network is caused to misclassify the perturbed input as a pre-selected target class $t$ and such that $r$ satisfies $\|r\|_\infty\leq\eta$. The algorithm progressively establishes the target universal perturbation via an iterative procedure over the data points in $X$. At each iteration, it computes a minimal perturbation $\Delta r_i$ that sends the current perturbed point $x_i + r_i$ toward the decision boundary of target class $t$ of the classifier, and then aggregates $\Delta r_i$ to the current instance of the target universal perturbation $r_i$, as illustrated in Fig.~\ref{fig:idea}(b). More specifically, as long as data point $x_i$ perturbed by the current $r_i$ is not classified as the target class $t$ by the attacked model, we solve the following optimization problem to find a supplemental $\Delta r_i$ that will lead to misclassification on $x_i$:
\begin{equation}
\begin{split} 
    \Delta r_i \leftarrow \mathop{\arg\min}_{\sigma} \|\sigma\|_\infty\\
    s.t.~\widehat{f}(x_i+r_i+\sigma)=t. 
    \label{4}
\end{split}
\end{equation}
We treat the problem in Eq.~\eqref{4} as a suitable optimization instance and solve it by existing optimization algorithms such as that given in \cite{carlini2017towards}.
To reduce the computational cost while ensuring that the constraint $\|r\|_\infty \leq \eta$ is satisfied, the updated perturbation $r$ is further clipped and projected onto the $\ell_\infty$ ball, with radius $\eta$ and centered at $0$, every $k$ iterations; the projection operator $\mathcal{P}_{\infty,\eta}$ is defined as follows:
\begin{equation}
\begin{split}
&\mathcal{P}_{\infty,\eta}=\mathop{\arg\min}_{r’} \|r-r’\|_2 \\
&s.t.~\|r’\|_\infty\leq\eta. 
\end{split}
\label{5}
\end{equation}
Then, we use the operator in Eq.~\eqref{5} to update the perturbation vector $r$ in the $i$th iteration as follows:
\begin{equation}
  r \leftarrow\left\{
                            \begin{array}{lr}
                            \mathcal{P}_{\infty,\eta}(r+\Delta r_i),~{\rm for}~i|k=0,i\neq0     &  \\
                            r+\Delta r_i,~{\rm otherwise }           &
                            \end{array}
                \right..
\end{equation}
When the attack success rate for target class $t$ exceeds the desired threshold $1-\delta$ on the perturbed dataset $X_r :=\{x_1+r,\ldots,x_s+r\} $, the algorithm is stopped. The success rate $ S_{ucc}(X_r)$ is defined as the likelihood of success that the perturbation will change the label to the target class $t$. In other words, the terminal condition of the algorithm is
\begin{equation}
   S_{ucc}(X_r) := \frac{1}{s}\sum_{i=1}^s1_{\widehat{f}(x_i + r) = t} \geq 1-\delta,
\end{equation}
where $1_{\widehat{f}(x_i + r) = t}$ is the indicator function.
The details of the algorithm are provided as Algorithm~\ref{alg:1}.

\begin{algorithm}
\setstretch{1}
\caption{Computation of targeted universal perturbation.}
\label{alg:1}
\LinesNumbered
\KwIn{Dataset $X$, classifier $\widehat{f}$, target class $t$, desired $L_\infty$-norm of the perturbation $\eta$, desired projection operator step size $k$, desired accuracy on perturbed data points $\delta$}
\KwOut{\textit{targeted-$(\delta,\eta)$-universal perturbation} (TUP) vector $r$}
Initialize $r\leftarrow 0$.

\While{$S_{ucc}(X_r) \textless 1-\delta$}{
    Shuffle the dataset $X$
    
    \For{every $x_i \in X$}{
        \If{$\widehat{f}(x_i+r) \neq t$}{
        \[\Delta r_i \leftarrow \mathop{\arg\min}_{\sigma}  \|\sigma\|_\infty~\]  \[s.t.\widehat{f}(x_i+r_i+\sigma)=t\] \eIf{$i|k=0~\rm{and}~i\neq 0$}{
            Update the perturbation using the projection operator:
            \[r \leftarrow \mathcal{P}_{\infty,\eta}(r + \Delta r_i)\]
            }{Update the perturbation:
        \[r \leftarrow r+\Delta r_i\]
        }
        }
    }
}
\end{algorithm}
\subsection{Region adversarial training}
\label{section:4.2}

\begin{algorithm}
\caption{Region adversarial training.}
\label{alg:2}
\LinesNumbered
\KwIn{labeled training data $D_{train} = \left \{ \left ( x_{i}, y_{i} \right ) \right \}_{i=1}^{N}$, target $t$ corresponding to the region where robustness is desired to be enhanced, the original pooly robust classifier $\widehat{f}$.}
\KwOut{a more robust model with parameter vector $\Theta^*$}
Initialize $\Theta^*$.

\While{not converged do}{
    \For{every $\left ( x_{i}, y_{i} \right ) \in D_{train}$}{
        \eIf{$\widehat{f}(x_i) \neq t$}{
        \[\left ( {x_{i}}^{\left ( 0 \right )},  {y_{i}}^{\left ( 0 \right )}\right )\leftarrow \left ( x_{i}, y_{i} \right )\]}{
        \[\left ( {x_{i}}^{\left ( t \right )},  {y_{i}}^{\left ( t \right )}\right )\leftarrow \left ( x_{i}, y_{i} \right )\]
        }
        randomly split $D_{train}^{(0)} = \left \{ \left ( {x_{i}}^{(0)}, {y_{i}}^{(0)} \right ) \right \}_{i=1}^{M}$ into $D_{train}^{(k)} = \left \{ \left ( {x_{i}}^{(k)}, {y_{i}}^{(k)}\right ) \right \}, k=\left \{ 1, 2 \right \}$ evenly

        \For{every $\left ( {x_{i}}^{\left ( 1 \right )},  {y_{i}}^{\left ( 1 \right )}\right )$
                   $\in D_{train}^{(1)} = \left \{ \left ( {x_{i}}^{(1)}, {y_{i}}^{(1)} \right ) \right \}_{i=1}^{n}$}{
        \[r_{i} = TUP\left ( {x_{i}}^{\left ( 1 \right )},  t \right )\]
        \tcp{Use Algorithm~\ref{alg:1} to get TUP perturbations}
        }}
        \[ J_{\rm{adv}}(\bm{\theta ^{*}},\bm{x+r_{t}},y) = \sum_{i=1}^{n}J\left (\bm{\theta ^{*}},\bm{{x_{i}}^{\left ( 1 \right )}+ r_{i}}, {y_{i}}^{\left ( 1 \right )} \right ) \]


        \begin{equation}\nonumber
        \begin{split}
        J_0(\bm{\theta^{*}},\bm{x},y) = & \sum_{i=1}^{N-M} J\left (\bm{\theta^{*}},\bm{{x_{i}}^{\left ( t \right )}}, {y_{i}}^{\left ( t \right)} \right )\\
        & + \sum_{i=1}^{M-n} J\left (\bm{\theta^{*}},\bm{{x_{i}}^{\left ( 2 \right )}}, {y_{i}}^{\left ( 2 \right )} \right )
        \end{split}
        \end{equation}

\[J = J_{0}(\bm{\theta ^{*}},\bm{x},y)+J_{\rm{adv}}(\bm{\theta ^{*}},\bm{x + r_{t}},y)\]
Apply $J$ to update model
}
\end{algorithm}

\textbf{Algorithm 2.} In order to use the TUP approach to enhance the adversarial robustness of deep networks, we introduce a training method, which we call region adversarial training (RAT). The purpose of the training is not to enhance the entire network in undifferentiated ways; instead, it focuses on the weaker regions of the network or the regions of most interest to the user. In region adversarial training, the network is not trained on all inputs from the training set perturbed but on a mixture of original training data and training data perturbed by the TUP method.
The targeted universal perturbation that is computed can be considered as containing more complex information of a certain class region's decision boundaries that the network has not yet learned from the original training set. The intuition behind region adversarial training is that incorporating this information into the training will improve the classification accuracy on adversarial examples of the classifier for this class. Formally, let $\Theta ^*$ be the weights of a neural network; then standard training learns $\Theta ^*$  as
\begin{equation}
    \Theta^* = \mathop{\arg\min}_{\theta} \mathbb{E}_{x\in\chi}J(\bm{\theta},\bm{x},y).
\end{equation}

The adversarial training proposed by Szegedy et al. \cite{szegedy2013intriguing} was originally for solving the following min--max formulation:
\begin{equation}
    \Theta^* = \mathop{\arg\min}_{\theta} \mathbb{E}_{x\in \chi}[\max\limits_{\delta \in \Delta(x)}J(\bm{\theta},\bm{x+\delta},y)], \label{9}
\end{equation}
where $\delta$ represents adversarial perturbations computed by some method; in \cite{szegedy2013intriguing}, a linear approximation method named Fast Gradient Sign Method  (FGSM) was used to generate $\delta$.
The original adversarial training process trained on the perturbed samples roughly, without direction or distinction. The region adversarial training method proposed here pays special attention to the region of the space where the classifier outputs a certain class label $t$ in the manifold represented by the network. Using this method, $\Theta ^*$ is computed as
\begin{equation}
\begin{split}
     \Theta^* =& \mathop{\arg\min}_{\theta} \mathbb{E}_{x\in \chi}[\max\limits_{\delta \in \Delta(x)}[J_0(\bm{\theta},\bm{x},y)\\
     &+J_{\rm{adv}}(\bm{\theta},\bm{x+\delta},y)]],\label{10}
\end{split}
\end{equation}
\begin{equation}
     J_0(\bm{\theta},\bm{x},y) = \sum_{x_i\in\chi,f(x_i)=t}J(\bm{\theta},\bm{x_i},y),\label{11}
\end{equation}
\begin{equation}
      J_{\rm{adv}}(\bm{\theta},\bm{x+\delta},y) = \sum_{x_i\in\chi,f(x_i)\neq t}J(\bm{\theta},\bm{x_i+\delta},y).\label{12}
\end{equation}
The saddle point problem in Eq.~\eqref{10} is similar to that in Eq.~\eqref{9} in its composition of an inner maximization problem and an outer minimization problem. The loss function $J_0$ in Eq.~\eqref{11} is independent of the perturbation $\delta$, and so the inner maximization problem in Eq.~\eqref{10} can be rewritten as
\begin{equation}
    J_0(\bm{\theta},\bm{x},y)+\max\limits_{\delta \in \Delta(x)}[J_{\rm{adv}}(\bm{\theta},\bm{x+\delta},y)].\label{13}
\end{equation}
Let $\bm{r_t}$ be the perturbation vector found by Algorithm~\ref{alg:1}; then $\bm{r_t}$ can be interpreted as a scheme for maximizing the loss $J_{\rm{adv}}$ in Eq.~\eqref{13}. Thus, the weights $\Theta ^*$ are computed by the region adversarial training as
\begin{equation}
     \Theta^* = \mathop{\arg\min}_{\theta} \mathbb{E}_{x\in \chi}[J_0(\bm{\theta},\bm{x},y)+J_{\rm{adv}}(\bm{\theta},\bm{x+r_t},y)].\label{14}
\end{equation}
Eq.~\eqref{14} can be used for any suitable loss function $J(\bm{\theta},\bm{x},y)$; in this paper, we use the common cross-entropy loss function for neural networks. The details of the algorithm are provided as Algorithm~\ref{alg:2}.  

The main overhead of RAT arises from computing TUPs. Lines $2$-$10$ in Algorithm~\ref{alg:1} compute a TUP vector until the attack success rate exceeds the threshold. The number of executions of lines $3$-$10$ in Algorithm~\ref{alg:1} depends on the choice of parameters $\delta$, and the attack success rate of the current perturbation, which is mainly affected by $k$ and $\eta$ . Choosing these parameters empirically allows these lines to be executed only once and accordingly, only one shuffle of the dataset $X$ is required. Lines $4$-$10$ in Algorithm~\ref{alg:1} update the TUP perturbation vector for shuffling the dataset $X$ once, and the time complexity is $O(n)$, where $n$ is the size of set $X$ that the TUP is computed on.

\section{Evaluation}
\label{section:5}
In this section, the cases for the experimental investigation are introduced. Before turning to our approach for generating adversarial examples and improving adversarial robustness, we describe the architectures of the models on which we evaluated the proposed approach and the datasets we used. Then, we describe how the TUPs were generated for MNIST and CIFAR-10, show the performance of our TUP attack, discuss the influence of parameter selection, and study the property of transferability across different models and the performance on small datasets. Finally, based on the experimental results, we discuss whether the proposed region adversarial training with TUPs can improve adversarial robustness not only against TUP itself but also against FGSM adversarial examples. Furthermore, we also remark on the size of the set $X$ needed to achieve the desired results.
\subsection{Experimental setup}
\label{section:5.1}
\textbf{Dataset description.} To ascertain the feasibility and effectiveness of the algorithm proposed in this paper, a series of experiments were performed on two widely used machine learning datasets, MNIST and CIFAR-10, which are commonly used to test the performance of numerous prevalent methods, such as those in \cite{madry2017towards}, \cite{carlini2017towards}, \cite{papernot2016limitations}, etc. We chose these datasets not only for the convenience of comparing the performance of our method with that of other methods, but also because the TUP method extracts the semantic information that is common to most training data, and experimenting on small-scale datasets can reduce the search space and improve the efficiency of the algorithms. The MNIST dataset is a collection of black and white images of handwritten digits; it contains $60,000$ $28\times 28$ training samples and $10,000$ test samples, each pixel of which is encoded as a real number between $0$ and $1$. The CIFAR-10 dataset consists of $60,000$ $32\times 32$ color images, which are divided into a training set of $50,000$ images and a test set of $10,000$ images, each pixel of which takes the value of a real number between $0$ and $255$ for three color channels. For both the MNIST and CIFAR-10 datasets, we created a validation set containing $5000$ examples from the training set. Each image in the MNIST and CIFAR-10 dataset is associated with a label from ten classes. In MNIST, the classes are the values ranging from $0$ to $9$, representing the digit written, and in CIFAR-10, the ten classes are airplane, automobile, bird, cat, deer, dog, frog, horse, ship, and truck.
\begin{table}[h]
\caption{Baseline accuracy (acc.) of five MNIST classifiers and five CIFAR-10 classifiers.}
\label{tabel:1}
\begin{tabular}{m{0.3\textwidth} m{0.1\textwidth}}
\hline
\textbf{Architecture (MNIST)}        & \textbf{Acc. (\%) }  \\ \hline
Classifier-M-Primary (Classifier$_{Mp}$)     & 99.34 \\
Classifier-M-Alternate-0 (Classifier$_{M0}$) & 99.31 \\
Classifier-M-Alternate-1 (Classifier$_{M1}$) & 99.38 \\
Classifier-M-Alternate-2 (Classifier$_{M2}$) & 99.30 \\
Classifier-M-Alternate-3 (Classifier$_{M3}$) & 99.35 \\ \hline
\textbf{Architecture (CIFAR-10)}     & \textbf{Acc. (\%) }   \\ \hline
Classifier-C-Primary (Classifier$_{Cp}$)     & 77.74 \\
Classifier-C-Alternate-0 (Classifier$_{C0}$) & 78.03 \\
Classifier-C-Alternate-1 (Classifier$_{C1}$) & 73.55 \\
Classifier-C-Alternate-2 (Classifier$_{C2}$) & 73.09 \\
Classifier-C-Alternate-3 (Classifier$_{C3}$) & 75.46 \\ \hline
\end{tabular}
\end{table}

\textbf{Architecture characteristics.} To begin our empirical explorations, we trained five networks each for the standard MNIST and CIFAR-10 classification tasks. The five networks differed only in their initial weights or their architectures. The baseline accuracies on clean data (unperturbed data) are listed in Table~\ref{tabel:1}. The details of the model architectures and the hyper-parameters we used are given in the Appendix. The performance of the networks on MNIST was comparable to state-of-the-art performance \cite{Cire2012Multi}, but note that the accuracy on CIFAR-10 was much lower for all five networks. The state-of-the-art accuracy on CIFAR-10 is higher \cite{Graham2014FractionalM}, but to achieve this performance, data augmentation or additional dropout must be used. In the context of adversarial robustness, researchers are typically concerned with the original data, and we achieved a test accuracy of $77.74\%$, which is very close to the state-of-the-art validation accuracy without any data augmentation \cite{Mairal2014Convolutional}. We did not attempt to increase this number through tuning hyper-parameters or any of the many other techniques available, as we wanted to use a typical convolution structure (based on the well-studied LeNet \cite{LeCun1999ObjectRW}) that is commonly used in other studies and training approaches that are identical to those presented in \cite{papernot2016distillation} and \cite{carlini2017towards} to make it easy for others to compare with or replicate our work.

\subsection{Crafting of adversarial examples using TUPs}
\label{section:5.2}
\begin{figure}
    \centering
    \includegraphics[width=3.2in]{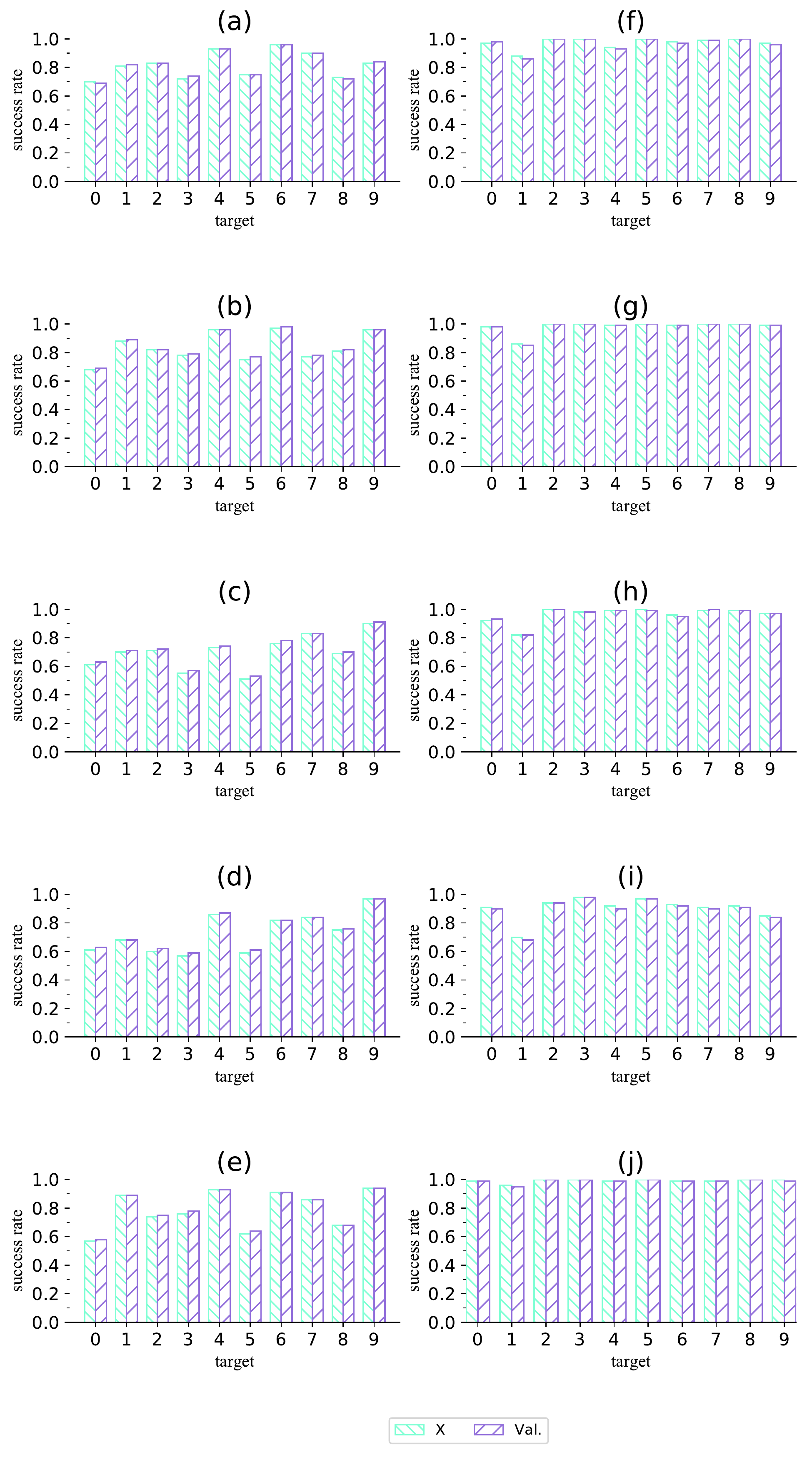}
    \caption{Success rates of TUP adversarial examples on $X$ and the disjoint validation set for targeted attacks of each target class (from 0 to 9). Left column: Success rate of attacks against the five networks on CIFAR-10; (a)--(e) correspond to models Classifier$_{Cp}$, Classifier$_{C0}$, Classifier$_{C1}$, Classifier$_{C2}$, and Classifier$_{C3}$, respectively. Right column: Success rate of attacks against the five networks on MNIST; (f)--(j) correspond to models Classifier$_{Mp}$, Classifier$_{M0}$, Classifier$_{M1}$, Classifier$_{M2}$, and Classifier$_{M3}$, respectively.}
    \label{fig:base_sr}
\end{figure}

\begin{figure*}
    \centering
    \includegraphics[width=\textwidth]{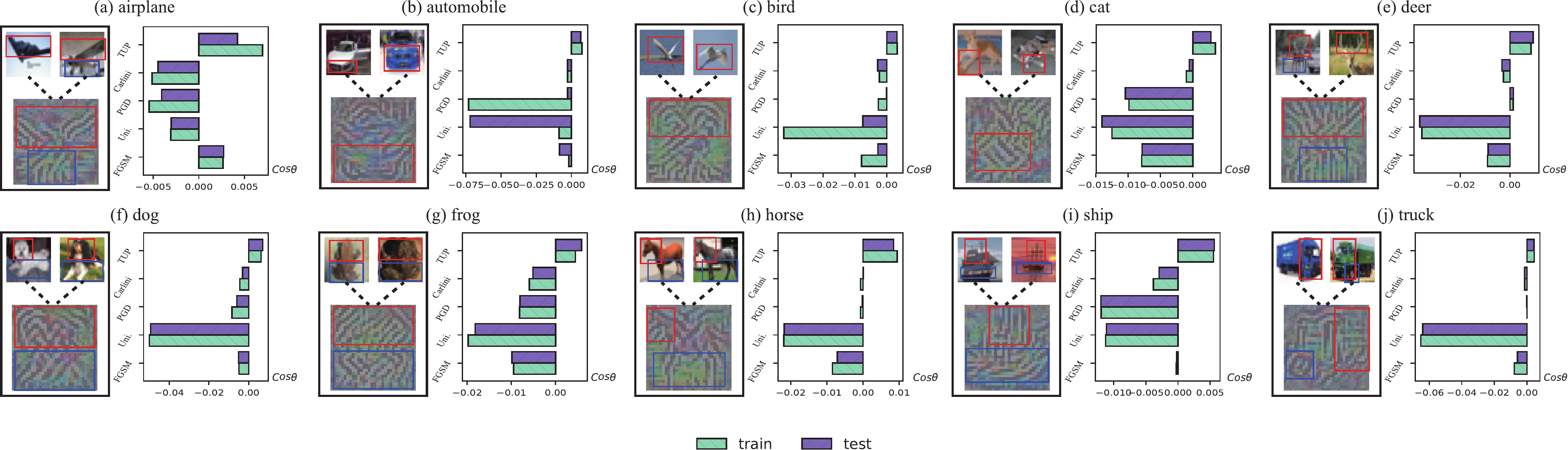}
    \caption{Visualization of semantic information contained in perturbations computed by TUP method for CIFAR-10. The ten classes (a)-(j) shown are the target classes chosen for the respective attacks. In each subgraph of (a)-(j), \emph{left}: The pixel values of the perturbations are scaled for visibility. To show the semantic information carried by the perturbation more clearly, two randomly selected images from the training set are displayed for each target class. Same-colored boxes on the perturbation images and the sample images indicate the same semantic concept. In each subgraph of (a)-(j), \emph{right}: The cosine similarity between the perturbations computed by the five adversarial example generation methods and the contour vector extracted from the randomly selected samples of the CIFAR-10 training set and test set.}
    \label{fig:perturbation}
\end{figure*}

\textbf{Success rate.} To evaluate the attack performance of the proposed algorithm, we report the success rate, which is defined as the proportion of samples that are misclassified as target class $t$ when perturbed by our perturbation, on CIFAR-10 and MNIST (Fig.~\ref{fig:base_sr}). For all of the model architectures, results are reported on set $X$, which was randomly selected from the training sets of CIFAR-10 and MNIST to compute the perturbation, and on a validation set that had never been used during the process of computing the perturbation. The set $X$ contained $10,000$ images, and the validation set contained $5000$ images for both CIFAR-10 and MNIST. As can be seen, the perturbations achieved quite high success rates for all sets of conditions, although there are some differences in the success rates because of differences in architecture, target classes, datasets, and parameter selection, which we discuss below.
Notably, these results demonstrate the universality property, namely, that any image in the validation set can be used to fool the classifier into misclassifying it as a target class $t$ (different from its source class) by the mere addition of the TUP perturbation computed on another disjoint set. Fig.~\ref{fig:adversarial} illustrates images before and after perturbation by TUPs; note that in most cases, the perturbations are nearly imperceptible. We display these perturbations in Fig.~\ref{fig:perturbation}, where the patterns of the perturbations are clearly shown and are seen to contain distinct semantic information.
\begin{figure}
    \centering
    \includegraphics[width=3.5in]{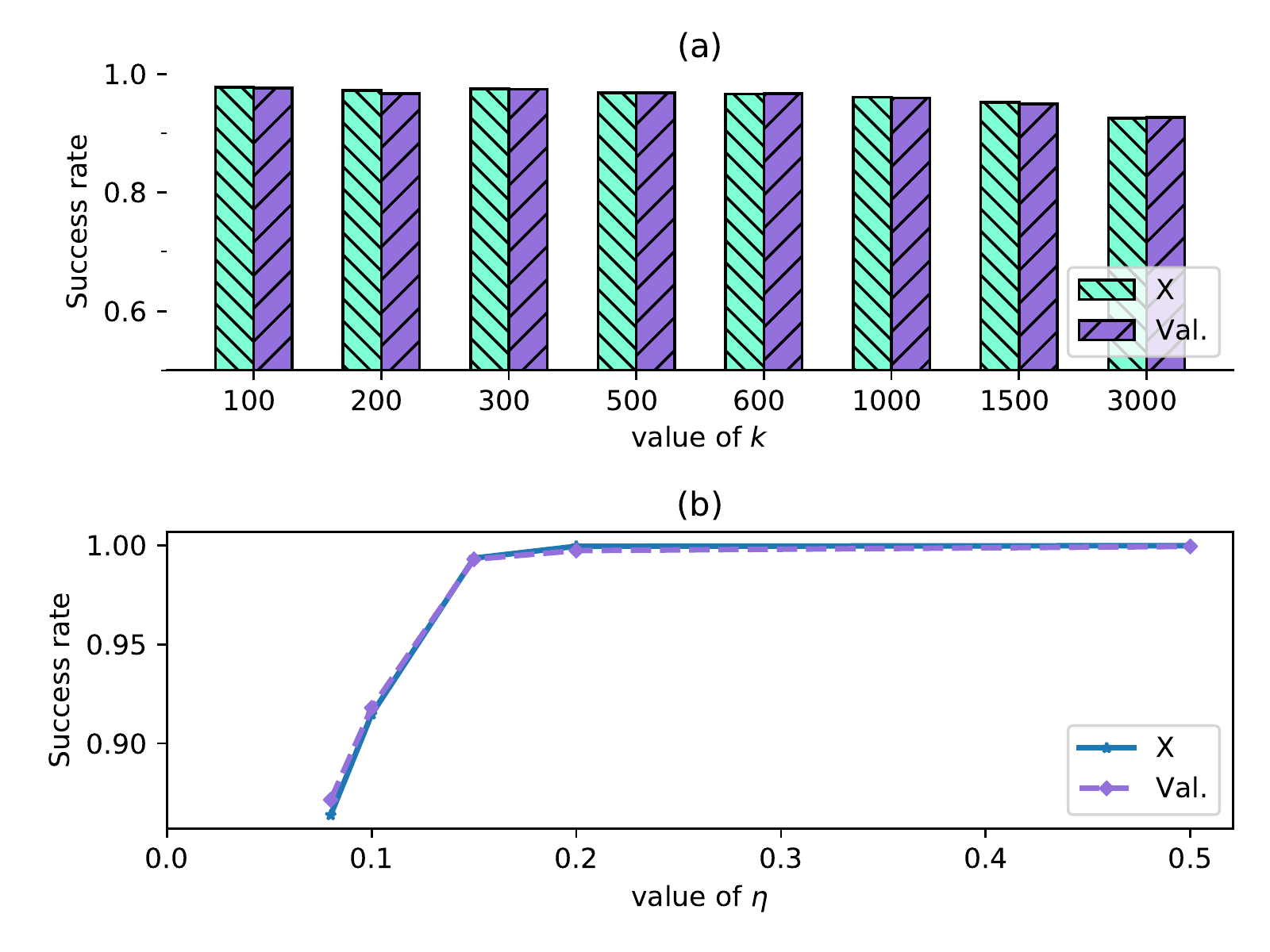}
    \caption{Effect of values of parameters $k$ and $\eta$ on attack success rate.}
    \label{fig:k_xi}
\end{figure}
Let $N$ be the number of images in set $X$, representing the size of $X$. In all of the above experiments, we used $k=N$, $\eta=0.1$ for CIFAR-10 and $k=N$, $\eta=0.8$ for MNIST, chosen empirically. Although these values for parameters $k$ and $\eta$ worked well enough, we explored further to learn whether there might be different options for other situations. The effect of the parameter values was evaluated on the baseline network Classifier$_{Cp}$, and some of these results are shown in Fig.~\ref{fig:k_xi}. Please note that in order to reduce the amount of calculation required, we chose a smaller set $X$, which included $3000$ CIFAR-10 images, to compute the adversarial perturbations and selected the target frog (class 6, chosen randomly from the ten classes) to use as an example.

Using a larger value for the projection step size $k$ results in fewer projection operations. Thus, it is natural to hypothesize that the success rate will decrease as $k$ increases. The results displayed in Fig.~\ref{fig:k_xi}(a) do not violate our intuition: With $k=100$, $97.61\%$ of the examples in the validation set disjoint with $X$ were classified incorrectly as the target class frog, whereas when $k$ increased to $3000$ (equal to the size of $X$) the attack success rate decreased to $92.66\%$. In contrast with this modest decrease in the attack success rate, it is surprising to see that the calculation time decreased dramatically as $k$ increased. When $k=3000$, it required only slightly more than half the time needed for $k=100$. Therefore, if an extremely high performance in terms of the success rate is not pursued, a larger value of $k$ is acceptable.

The effect of parameter $\eta$, the radius of the $l_\infty$ ball on which the perturbation is projected during the computation, is rather interesting. We varied $\eta$ from $0.08$ to $0.5$ and found that the success rate increased linearly from $\eta = 0.08$ to $\eta = 0.15$ and then plateaued from $\eta = 0.15$ to $\eta = 0.5$; clearly, therefore, increasing $\eta$ increased the attack success rate of a TUP perturbation, as displayed in Fig.~\ref{fig:k_xi}(b). It should be noted that the method proposed in Algorithm~\ref{alg:1} is not theoretically guaranteed to converge to the optimal solution, as it operates in a greedy way. When $\eta$ was chosen to be very small, the success rate oscillated back and forth far below the desired performance, and we observed that the smaller the value, the more violent the oscillation, and thus the more difficult the convergence.

To qualitatively and quantitatively study whether the perturbations generated by our TUP method contain the correct semantic information of the target classes, we visualize the perturbations corresponding to each of the ten classes of CIFAR-10 and compare them with images randomly selected from the training set. In Fig.~\ref{fig:perturbation}, the patterns of the perturbations are clearly shown; moreover, they contain distinct semantic information. In addition, we randomly selected $1000$ images from the training set and test set for each class and used the Canny \cite{canny1986computational} detector to detect the edges in these images; then we  calculated the cosine similarity between the edge vectors and the perturbation vectors. The average values of the cosine similarities corresponding to the $1000$ training images and test images for TUP, C\&W, Uni., PGD and FGSM are also reported in Fig.~\ref{fig:perturbation}. Note that for all the ten target classes, the average value of cosine similarity corresponding to our TUP method is the closest to $1$; this means the perturbations computed by our TUP method are more similar to the edge vectors (which often carry important semantic information \cite{szeliski2010computer}) of the original images belonging to the target class. There are numerous algorithms for semantic segmentation, and the effect of extracting semantic information using TUPs can be further studied by considering these algorithms. This is an important research avenue that we reserve for future work.

\begin{figure}
    \centering
    \includegraphics[width=3.5in]{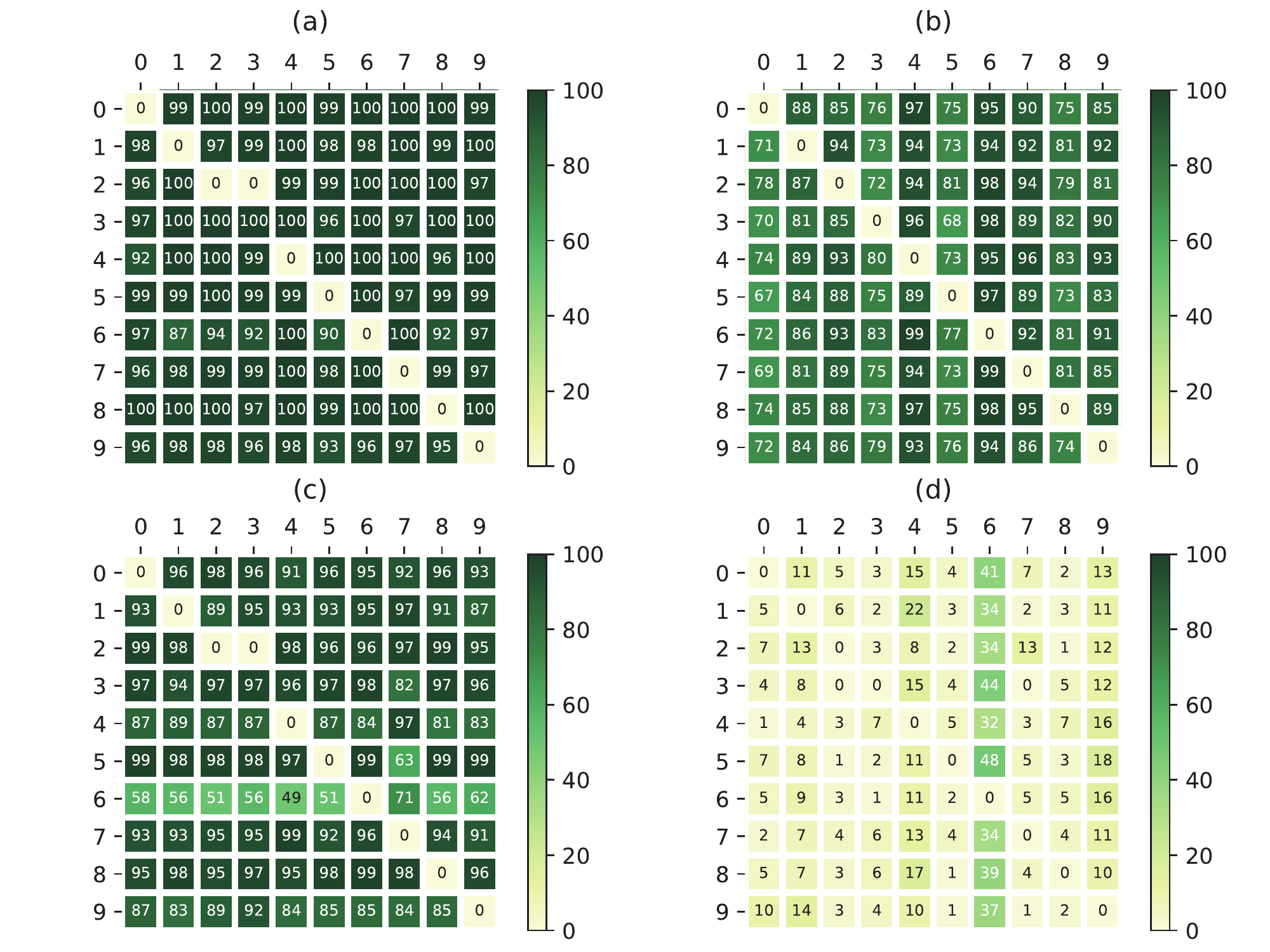}
    \caption{Heat maps of the number of times an attack was successful with the corresponding source--target class pair, for both targeted and untargeted attacks by TUP and FGSM. (a) TUP untargeted attacks, (b) TUP targeted attacks, (c) FGSM untargeted attacks, and (d) FGSM targeted attacks.}
    \label{fig:heatmap}
\end{figure}

We compared the proposed TUP method with the most well-known version of FGSM on the baseline model Classifier$_{Cp}$. We used Cleverhans \cite{papernot2018cleverhans} to re-implement the ``target class'' variation \cite{kurakin2016adversarial} of FGSM, as the TUP method can be used to launch a targeted attack. We generated $100$ TUP adversarial examples and $100$ FGSM adversarial examples for each source--target pair on CIFAR-10. In Fig.~\ref{fig:heatmap}, the left column represents the number of successful untargeted attacks out of the $100$ attacks for each source--target pair, and the right column represents the number of targeted attacks. The first row corresponds to the TUP attack, and the second row to the FGSM attack. As shown by the heat maps, the TUP method had high success rates in both targeted and untargeted attacks, whereas FGSM only achieved a comparable success rate in the untargeted attacks, performing poorly in the targeted attacks. The number of successful TUP attacks was almost evenly distributed across each source--target class pair, and the heat maps for TUP are almost symmetric. This means that for two classes A and B, perturbing images from A to B is approximately as difficult as perturbing from B to A for a TUP attack. For an FGSM attack, however, there exist some specific source--target class pairs that are much more vulnerable than others in both targeted and untargeted attacks. This indicates that the TUP method has found a universal way to perturb the inputs in a certain direction as specified by the target class, whereas FGSM is inclined to perturb the original images in the direction of some vulnerable target class shared by many data points.

\textbf{Cross-model transferability.} Previous work demonstrated the transferability property of adversarial examples, that is, that adversarial examples crafted to mislead one model can affect other models provided they are trained to perform the same task, even if their architectures are different or their training sets are disjoint \cite{szegedy2013intriguing}, \cite{papernot2016transferability}, \cite{papernot2017practical}. To measure the cross-model transferability of perturbations crafted by the TUP method, i.e., the extent to which the perturbations computed for a specific architecture are effective for another, we computed perturbations for each architecture on both MNIST and CIFAR-10 for each target class and fed the addition of each universal perturbation to the other network. We report the average attack success rate for the ten target classes on all other architectures for the same dataset in Table~\ref{tabel:3}. The perturbations had an average cross-model success rate of greater than $63\%$ on MNIST and $41\%$ on CIFAR-10. In the best cases, perturbations computed for the Classifier$_{M0}$ network had a success rate of $81.03\%$ with Classifier$_{M3}$ (on MNIST), and perturbations computed for Classifier$_{Cp}$ had a $62.97\%$ success rate with Classifier$_{C0}$ (on CIFAR-10). We observed that the perturbations computed for different architectures had discrepant transfer capabilities across other architectures. For example, the perturbations computed for Classifier$_{M1}$ (all above $56\%$, and best case $77.92\%$) and Classifier$_{Cp}$ (all above $40\%$ except for Classifier$_{C1}$, and best case $62.97\%$) generalized better than other architectures on the same dataset. The bold numbers in Table~\ref{tabel:3} represent the highest cross-model success rate of the TUPs calculated on each model. These results show that the TUPs we created do transfer to some extent across models, thereby demonstrating that our TUP perturbations are not an artifact of a specific network nor of a particular selection of training set but have a degree of universality with respect to both data points and architectures.

\begin{table}
\caption{Cross-model success rates (\%) on CIFAR-10 and MNIST. Rows indicate the architecture for which the TUPs were computed, and columns indicate the architecture for which the success rate is reported. The maximum value in each row is shown in bold font.}
\label{tabel:3}
\small
\scalebox{0.7}{
\begin{tabular}{cccccc}
\toprule
       & \textbf{Classifier$_{Cp}$} &
       \textbf{Classifier$_{C0}$} &
       \textbf{Classifier$_{C1}$}&
       \textbf{Classifier$_{C2}$} &
       \textbf{Classifier$_{C3}$} \\ \hline

\textbf{Classifier$_{Cp}$}  & - & \textbf{62.97}          & 32.50          & 40.60          & 52.64          \\ \hline
\textbf{Classifier$_{C0}$} & \textbf{45.75 }         & - & 27.64          & 36.88          & 45.23          \\ \hline
\textbf{Classifier$_{C1}$} & 49.77          & \textbf{52.28}          & -  & 37.95          & 43.92          \\ \hline
\textbf{Classifier$_{C2}$} & 38.20          & \textbf{41.90}          & 25.64          & - & 38.11          \\ \hline
\textbf{Classifier$_{C3}$} & 47.22          & \textbf{52.71}          & 26.56          & 39.50          & - \\ \hline
\hline
       & \textbf{Classifier$_{Mp}$}&
       \textbf{Classifier$_{M0}$}&
       \textbf{Classifier$_{M1}$}&
       \textbf{Classifier$_{M2}$}&
       \textbf{Classifier$_{M3}$}       \\ \hline
\textbf{Classifier$_{Mp}$} & - & 78.78          & 51.29          & 35.75          & \textbf{79.10}          \\ \hline
\textbf{Classifier$_{M0}$} & 69.01          & - & 57.19          & 37.59          & \textbf{81.03 }         \\ \hline
\textbf{Classifier$_{M1}$} & 70.81          & 76.27          & - & 56.94          & \textbf{77.92}          \\ \hline
\textbf{Classifier$_{M2}$} & 63.52          & \textbf{68.30}         & 61.98          & - & 67.48          \\ \hline
\textbf{Classifier$_{M3}$} & 68.91          & \textbf{77.66}          & 51.92          & 36.20          & - \\
\bottomrule
\end{tabular}}
\end{table}
\textbf{Size of set $X$.} As described previously, each of the TUPs above was computed for a set $X$, a random selection of $10,000$ examples from the training set (excluding images that were originally classified as class $t$). Is such a large set $X$ necessary to achieve similar attack success rates? The answer to this question may allow the TUP method to be made more practical. Using a smaller set $X$ does allow a more realistic assumption regarding the attacker's access to data; that is, that the attacker has access to only a subset of the training data rather than full access to any examples that were used in training the target model. Meanwhile, using a smaller set $X$ makes the algorithm faster.

\begin{table}
\caption{Success rates (\%) corresponding to different sizes $N$ for set $X$ on MNIST and CIFAR-10.}
\label{tabel:4}
\small
 \resizebox{0.5\textwidth}{12mm}{
\begin{tabular}{cccccccccc}
\toprule
\multirow{2}*{\textbf{Dataset}} & \multicolumn{9}{c}{$N$} \\\cline{2-10}
&100& 200& 300& 400& 500& 600& 700& 800& 900\\
\midrule
MNIST    & 81.79 & 83.57 & 89.16 & 92.02 & 93.32 & 95.06 & 95.36 & 96.08 & 96.08 \\ \hline
CIFAR-10 & 69.37 & 88.88 & 92.45 & 94.93 & 95.09 & 96.69 & 97.54 & 97.90 & 98.63 \\ \hline
\multirow{2}*{\textbf{Datasets}} & \multicolumn{9}{c}{$N$} \\\cline{2-10}
& 1000& 2000& 3000& 4000& 5000& 6000& 7000& 8000& 9000\\
\midrule
MNIST    & 96.34 & 96.61 & 96.63 & 97.20 & 97.47 & 98.22 & 98.61 & 98.62 & 99.26 \\ \hline
CIFAR-10 & 98.67 & 99.61 & 99.79 & 99.83 & 99.85 & 99.90 & 99.90 & 99.93 & 99.94 \\ \bottomrule
\end{tabular}}
\end{table}

Table~\ref{tabel:4} shows the success rates on the validation sets created with TUPs computed on variously sized subsets of the training set. We repeated the experiment ten times for each set $X$, each time randomly selecting the attack target $t$ from the ten classes of CIFAR-10 and MNIST; we report the average results for the ten trials for each $X$. To eliminate the effect of different projection steps and focus on the influence of the size of set $X$, the projection operator was omitted here; although this does cause the success rate to be higher than was shown before (at the expense of the quality of perturbed images), it does not affect the trend of the change in success rate as the size of $X$ is varied.

We might expect that perturbations computed on higher numbers of data samples to result in higher attack success rates, and this holds true when set $X$ contains fewer than $1000$ samples. With a set $X$ containing just $100$ CIFAR-10 images, the attack was successful for $69.37\%$ of the images on the validation set, and when perturbations were computed on $100$ MNIST images, the attack succeeded in more than $80\%$ of cases. When set $X$ was expanded to contain $1000$ samples, the perturbation computed on $X$ fooled $98.67\%$ of the validation images on CIFAR-10 and $96.34\%$ on MNIST; the success rates did not change much after that. This surprising result suggests that the proposed method is able to extract a large amount of useful information from a very small dataset.
\begin{figure}
\centering
\includegraphics[width=3.2in]{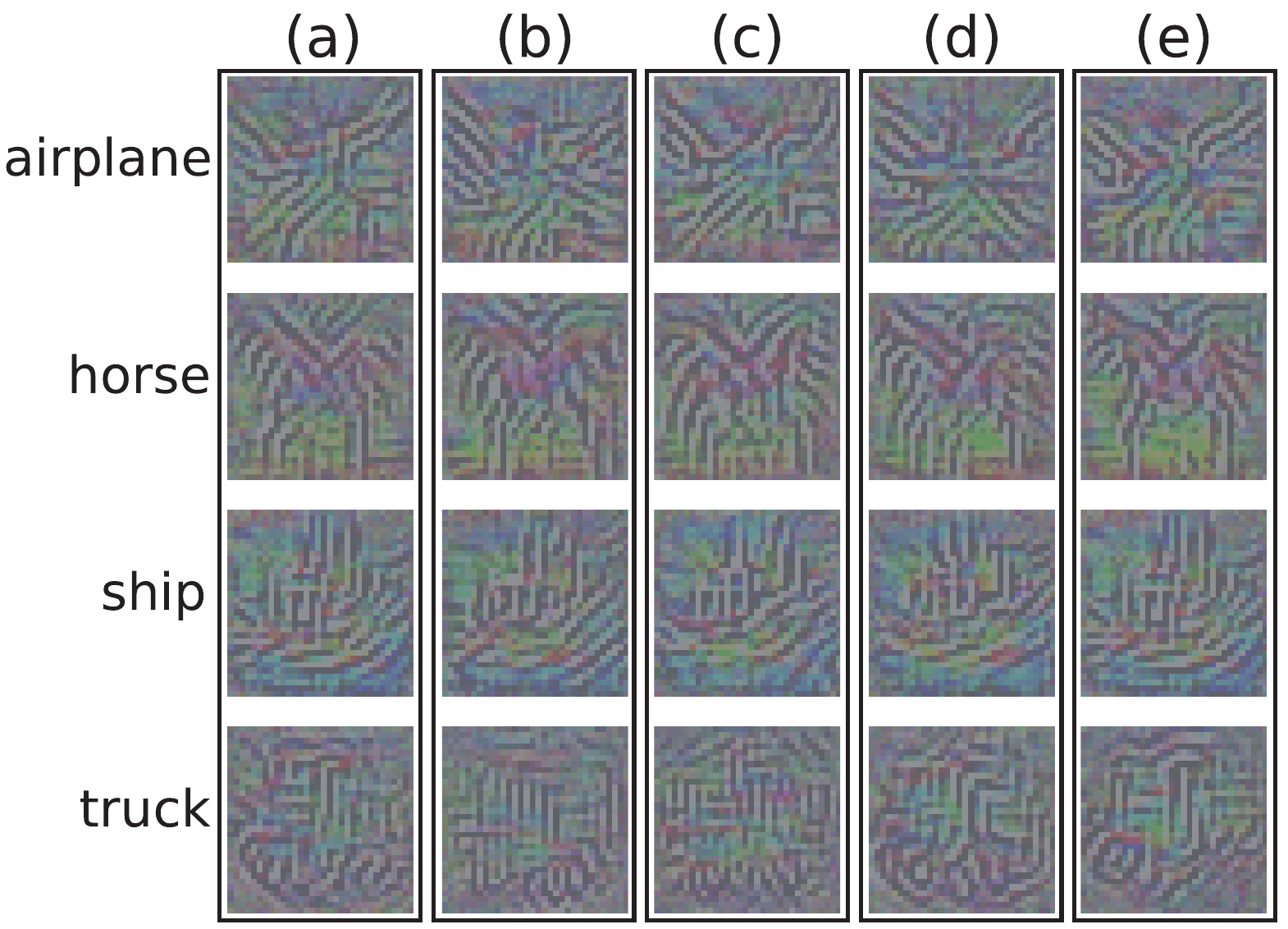}
\caption{Some TUP perturbations generated on sets $X$ of different sizes, using four randomly selected classes as examples. Columns (a)--(e) correspond to $N=1000$, $3000$, $5000$, $7000$, and $9000$, respectively.}
\label{fig:tmp2}
\end{figure}

To illustrate this observation, we display the perturbations of four randomly selected target classes on CIFAR-10 corresponding to different sizes of set $X$ in Fig.~\ref{fig:tmp2}. As the images show, explicit and rich semantic information was captured by perturbations computed on a very small $X$; the perturbations differed only slightly when computed on $X$ sets of different sizes. This hints that the structure of the dataset is quite meaningful in the construction of TUPs, whereas the quantity of data has no sizable effect.

\subsection{Effect of region adversarial training on adversarial robustness}
\begin{figure*}
\centering
\includegraphics[scale=0.6]{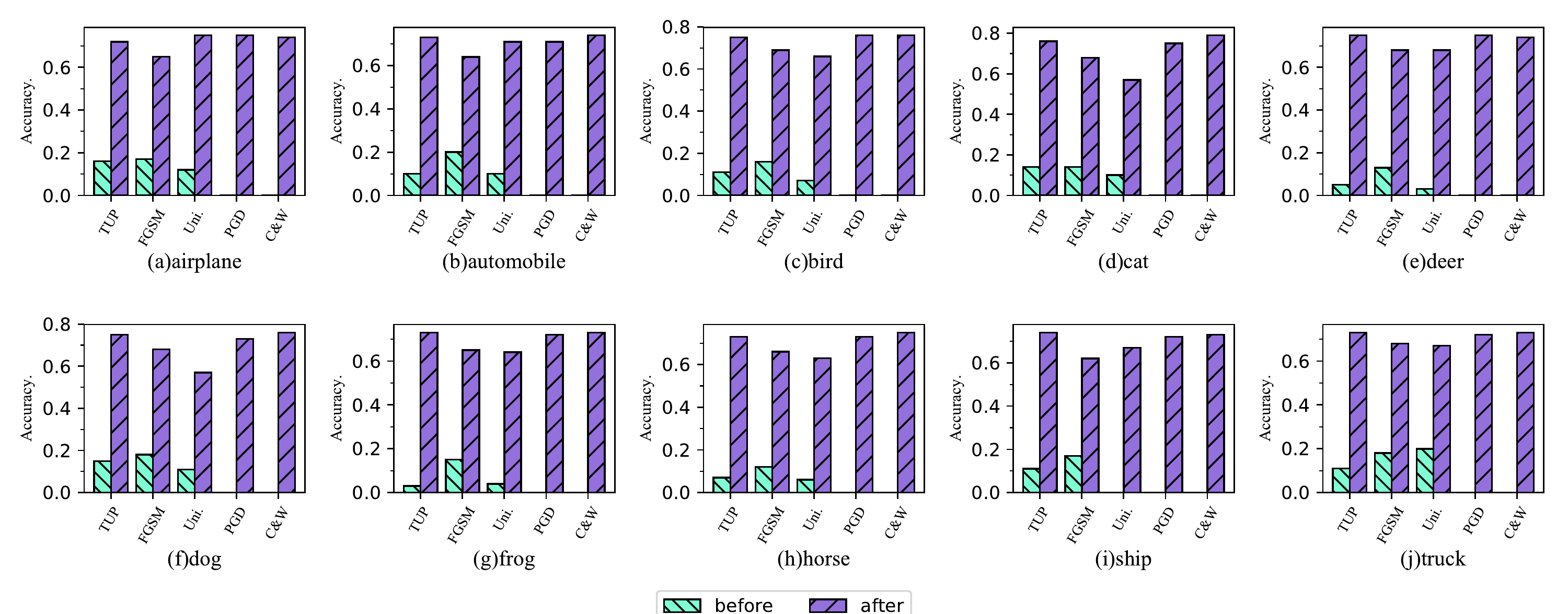}
\caption{Comparison of accuracy before and after region adversarial training based on TUP against TUP, FGSM, Uni., PGD, C\&W attacks. }
\label{fig:7}
\end{figure*}

We now examine the effect of region adversarial training with perturbed examples on the baseline models Classifier$_{Cp}$ and Classifier$_{Mp}$. We used the TUP perturbations computed for the networks Classifier$_{Cp}$ and Classifier$_{Mp}$ (described in Section~\ref{section:5.2} and presented in Fig.~\ref{fig:base_sr}) and performed region adversarial training according to the method described in Section~\ref{section:4.2}. In the RAT process, the TUP computed for one target class was used at a time, and we performed experiments on the ten target classes separately. Specifically, we included the adversarial counterparts of the original data during training through the simple addition of a targeted TUP perturbation to all of the clean examples classified as classes other than the target class $t$ by the attacked network. Then, we retrained the two baseline models for $50$ epochs.  We report the classification accuracy for the perturbed adversarial examples on the test set in Fig.~\ref{fig:7}. We observe that although the accuracy was not as high as that attained on the clean dataset, the use of region adversarial training did greatly improve the classification accuracy on adversarial examples compared with the accuracy before retraining. As Fig.~\ref{fig:7} shows, the accuracy on the perturbed test set rose to more than $72\%$ for each class in CIFAR-10; the previous accuracy was less than $31\%$ for all classes (average $10.34\%$, minimum $3.18\%$). For MNIST, the accuracy increased to over $98\%$ for each class, compared with less than $14\%$ (average $3.09\%$, minimum $0.07\%$) before. Note that in all of experiments reported in this paper, better results were obtained on MNIST than on CIFAR-10. One key reason is that the models we trained in this study perform better on the clean MNIST dataset than on the clean CIFAR-10 (as explained in Section~\ref{section:5.1}); thus, we could say that the model Classifier$_{Mp}$ is more powerful than Classifier$_{Cp}$ when they are performing their respective tasks. On the other hand, the MNIST dataset contains only black and white images, which have a pure background. In addition, in order to provide heuristic comparisons, we were more conservative in the parameter selections for CIFAR-10.

\begin{table*}
\centering
\caption{Accuracy(\%) on test set against TUP and FGSM adversarial examples corresponding to different sizes of set $X$ for the ten target classes.}
\small
\resizebox{\textwidth}{!}{
\begin{tabular}{ccccccccccccccccccccc}
\toprule
\multirow{2}*{\textbf{$N$}}
     & \multicolumn{2}{c}{0} & \multicolumn{2}{c}{1} & \multicolumn{2}{c}{2} & \multicolumn{2}{c}{3} & \multicolumn{2}{c}{4} & \multicolumn{2}{c}{5} & \multicolumn{2}{c}{6} & \multicolumn{2}{c}{7} & \multicolumn{2}{c}{8} & \multicolumn{2}{c}{9} \\
     \cline{2-21}
     & TUP & FGSM & TUP & FGSM& TUP & FGSM & TUP& FGSM & TUP & FGSM & TUP & FGSM & TUP & FGSM& TUP& FGSM& TUP & FGSM & TUP & FGSM \\ \hline
1000 & 24.43     & 17.20      & 24.39     & 20.00      & 27.49     & 15.74      & 27.67     & 14.16      & 15.21     & 12.69      & 20.69     & 17.50      & 15.62     & 14.78      & 24.96     & 12.03      & 26.27     & 16.86      & 18.53     & 18.33      \\ \hline
2000 & 23.39     & 17.20      & 22.31     & 20.00      & 25.47     & 15.74      & 24.66     & 14.16      & 13.99     & 12.69      & 18.60     & 17.50      & 14.04     & 14.78      & 21.21     & 12.03      & 25.50     & 16.86      & 17.88     & 18.33      \\ \hline
3000 & 23.13     & 17.20      & 22.18     & 20.00      & 23.19     & 15.74      & 23.73     & 14.16      & 10.26     & 12.69      & 17.68     & 17.50      & 10.81     & 14.78      & 11.46     & 12.03      & 22.47     & 16.86      & 15.09     & 18.33      \\ \hline
4000 & 22.31     & 17.20      & 19.93     & 20.00      & 22.12     & 15.74      & 19.30     & 14.16      & 8.70      & 12.69      & 16.19     & 17.50      & 7.02      & 14.78      & 10.74     & 12.03      & 21.69     & 16.86      & 12.68     & 18.33      \\ \hline
5000 & 22.18     & 17.20      & 19.64     & 20.00      & 22.18     & 15.74      & 16.11     & 14.16      & 7.23      & 12.69      & 15.74     & 17.50      & 6.70      & 14.78      & 9.61      & 12.03      & 20.49     & 16.86      & 12.01     & 18.33      \\ \hline
6000 & 19.93     & 17.20      & 17.81     & 20.00      & 19.91     & 15.74      & 15.19     & 14.16      & 6.56      & 12.69      & 15.34     & 17.50      & 5.70      & 14.78      & 8.36      & 12.03      & 19.37     & 16.86      & 11.59     & 18.33      \\ \hline
7000 & 19.64     & 17.20      & 16.98     & 20.00      & 15.56     & 15.74      & 14.57     & 14.16      & 6.36      & 12.69      & 14.57     & 17.50      & 4.94      & 14.78      & 8.22      & 12.03      & 18.90     & 16.86      & 8.96      & 18.33      \\ \hline
8000 & 17.81     & 17.20      & 14.43     & 20.00      & 15.49     & 15.74      & 14.26     & 14.16      & 6.01      & 12.69      & 13.59     & 17.50      & 4.63      & 14.78      & 6.74      & 12.03      & 17.72     & 16.86      & 8.02      & 18.33      \\ \hline
9000 & 16.98     & 17.20      & 14.13     & 20.00      & 13.94     & 15.74      & 12.87     & 14.16      & 5.92      & 12.69      & 13.13     & 17.50      & 4.16      & 14.78      & 6.53      & 12.03      & 16.34     & 16.86      & 6.98      & 18.33     \\ \bottomrule
\end{tabular}}
\label{table:5}
\end{table*}

\begin{figure*}
\centering
\includegraphics[scale=0.6]{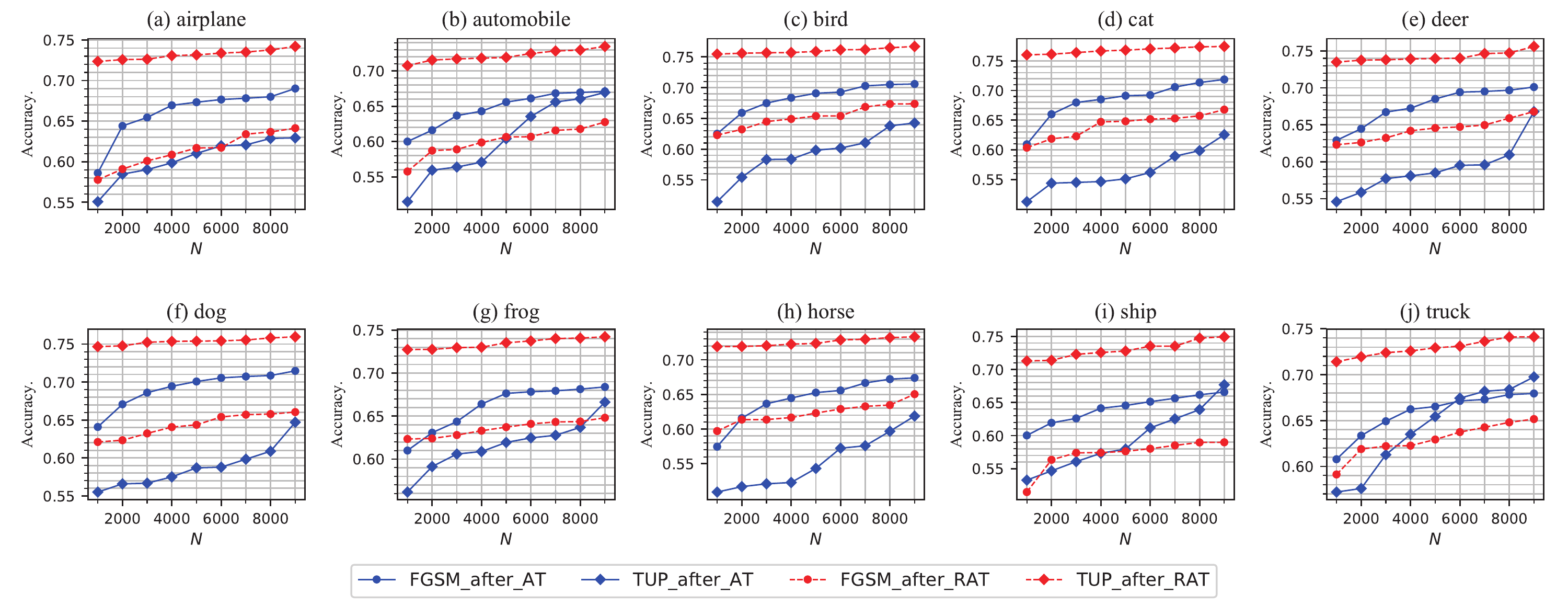}
\caption{Accuracy against TUP and FGSM attacks on test set before and after region adversarial training based on TUP (red) or classical adversarial training based on FGSM perturbations (blue). }
\label{fig:8}
\end{figure*}

Another finding is that region adversarial training using the TUPs not only strengthens the adversarial robustness to TUP perturbations themselves but is also effective against other adversarial attacks, such as the most well-known attack methods, FGSM \cite{goodfellow2014explaining}, Uni. \cite{moosavi2017universal}, PGD \cite{madry2017towards} and C\&W \cite{carlini2017towards}. We generated $1000$ targeted adversarial examples corresponding to each of FGSM, C\&W, and PGD for all classes in the CIFAR-10 test sets. For the untargeted method Uni., we generated a total of $10,000$ adversarial examples and calculated the accuracy based on the actual misclassifications of the attack examples. Note that we launched the PGD and C\&W attacks excessively harshly, similar to \cite{madry2017towards} and \cite{carlini2017towards}; that is, we altered each adversarial example until the attack was successful. This is a more extreme case that represents a stronger attack capability; consequently, the classification accuracy on the clean test dataset is zero for PGD and C\&W in Fig.~\ref{fig:7}. The results show that the networks trained using region adversarial training exhibited greater robustness properties that were not limited to the perturbations seen by the models during retraining; their ability to correctly classify unseen patterns, such as those corresponding to FGSM, Uni., PGD and C\&W, was also greatly improved, even under extreme attacks.

One question that remains is the following: Given that a TUP attack on a very small set can be quite powerful (as we have demonstrated), can region adversarial training with such attacks still improve robustness further? To investigate this issue, we measured the accuracies on the test set of CIFAR-10 against TUP and FGSM attacks after region adversarial training with TUPs computed on $X$ sets of different sizes for Classifier$_{Cp}$; these are reported in Fig.~\ref{fig:8}. For comparison, the results after adversarial training \cite{goodfellow2014explaining} with FGSM are also shown in Fig.~\ref{fig:8}. The original accuracies (before retraining) are shown in Table~\ref{table:5}. Note that only the number of images (from the training set) needed to generate adversarial examples for adversarial training has been changed; the final accuracy was calculated on the test set. As FGSM computes perturbations on a single image
at a time, whereas TUP computes an image-agnostic perturbation and then simply adds the perturbation to the clean input, the accuracies for FGSM shown in Table~\ref{table:5} do not change with $N$.

From the results shown, we find that region adversarial training based on the TUP algorithm offers comparative advantages in improving adversarial robustness through heuristics-based techniques. Firstly, both region adversarial training (RAT) based on TUP and adversarial training (AT) based on FGSM improve the test accuracy of the adversarial patterns they used during the retraining. In all cases of the ten target classes, however, RAT improves the accuracy of TUPs more than AT improves the accuracy of FGSM. Secondly, RAT also improves the robustness of the network against FGSM; in fact, there is not much of a gap between RAT and AT in their improvement of FGSM accuracy. By contrast, AT improves the accuracy of TUP much less than does RAT. Thirdly, the accuracy--$N$ curves for RAT are flatter than those for AT on both TUP and FGSM; this indicates that only a small number of samples are needed for the RAT method to achieve good results in enhancing adversarial robustness. This might be because of the difference in the principles of the two methods, AT hoping that the network can extract omitted information from a large number of adversarial examples on its own during retraining, and RAT using the missing semantic information near the classification boundary to guide the network's training.

\section{Conclusion}
\label{section:6}
. In summary, the method proposed in this paper improves the adversarial robustness of deep neural networks by emphasizing to deep models the missed semantic information of the region around the decision boundary. Our research builds on recent research on the generation of image-agnostic universal adversarial perturbations to fool deep neural networks, but it does so with attention to two entirely different goals: to have the perturbations extract the unlearned semantic information of a specific region in the manifold represented by a network and to use them to enhance the robustness of the network. We have proposed an algorithm named TUP to extract this information that the model has not yet learned but that is essential for correctly classifying the adversarial examples. The algorithm uses an iterative process on a subset of the training set to obtain a universal property across inputs, as many previous algorithms have done, but we interfered with the iterative process to push it toward the region corresponding to a specified target class. Furthermore, to enhance adversarial robustness, we designed region adversarial training based on the TUP perturbations. Experimental results on two datasets and ten classifiers show that region adversarial training based on the TUP algorithm not only improves robustness against TUPs but also markedly improves robustness against FGSM, Uni., PGD, and C\&W perturbations.

The TUP algorithm uses just a few training samples to effectively extract the semantic information obscured by the blind spots of the deep models, and at the same time it provides a powerful adversarial attack method that exhibits transferability across different architectures. The proposed region adversarial training method based on the TUP algorithm offers an efficient way to enhance the robustness of classifiers, especially the robustness of the region corresponding to a specific class, as the perturbation is universal for each class. By simply calculating a TUP perturbation on a very small set and then adding the perturbation to clean images, the method obtains the adversarial examples required for region adversarial training. The proposed approach provides new ideas for enhancing the adversarial robustness of DNNs and can be used as a fast and efficient tool. In our future work, we plan to explore the possibility of using multiple target classes concurrently during RAT and the method of selecting these target classes. Research on the methodology of selecting a few classes from among all the classes will provide insights on the geometric correlations between different parts of the decision boundary. Moreover, such research may also help improve the efficiency of the RAT algorithm while further enhancing the robustness of models, enabling the proposed technique to be scaled to large datasets such as ImageNet \cite{deng2009imagenet}. In our future works, we will perform more detailed investigations in this regard. Finally, as DNNs are increasingly leveraged to improve the accuracy of many security-sensitive applications, such as biometric systems, it provides our algorithm with promising application scenarios ranging from cellphone authentication to airport security systems. An interesting potential research involves how to apply the algorithm proposed in this paper to these systems, which requires one step forward and more effort. For example, how to make our algorithm more effective to extract useful semantic information from biometric features which are more detailed (such as facial features, palmprints, ears and irises), and how to make the algorithm work efficiently in complex systems are worthy of further investigation.

\section*{Acknowledgment}
This work was supported by the National Natural Science Foundation of China (Grant Nos.  U19A2081, 61802270) and the Fundamental Research Funds for the Central Universities (Grant No. 2019SCU12069, SCU2018D018).



%





\ifCLASSOPTIONcaptionsoff
  \newpage
\fi





\bibliographystyle{IEEEtran}
\bibliography{Bibliography}

\vfill


\end{document}